\pdfoutput=1
\documentclass[10pt, a4paper]{article}

\usepackage{lrec-coling2024} %

\usepackage{xstring}
\usepackage{color}
\usepackage{xspace,mfirstuc,tabulary}
\usepackage{inconsolata}
\usepackage{amsmath}
\usepackage{booktabs}
\usepackage{nccmath}
\usepackage{amssymb}
\usepackage{graphicx}
\usepackage{mathrsfs}
\usepackage{multirow}
\usepackage{xcolor}
\usepackage{tikz}
\usepackage{tikz-qtree}
\usepackage{tikz-qtree-compat}
\usepackage{tikz-dependency}
\usepackage{url}
\usepackage{caption}
\usepackage{changepage}
\usepackage{microtype}
\usepackage{forest}
\usepackage{subfigure}
\usepackage{mathtools}
\usepackage{algorithm}
\usepackage{algorithmic}
\usepackage{anyfontsize}
\usepackage{arydshln}
\usepackage{dashrule}
\usepackage{fontawesome}
\usetikzlibrary {arrows.meta}
\definecolor{midnightblue}{HTML}{005c7f}
\definecolor{brickred}{HTML}{b92622}
\definecolor{munsell}{rgb}{0.0, 0.5, 0.0}

\newcommand{\suda}{\textsuperscript{\faStarO}}
\newcommand{\huawei}{\textsuperscript{\faMoonO}}

\title{High-order Joint Constituency and Dependency Parsing}

\name{
    Yanggan Gu\suda,
    Yang Hou\suda,
    Zhefeng Wang\huawei,
    Xinyu Duan\huawei,
    Zhenghua Li\suda\sthanks{$^*$Corresponding author.}
} 

\address{
    \suda School of Computer Science and Technology, Soochow University, China \\
    \huawei Huawei Inc., China \\
    yanggangu@outlook.com,
    yhou1@stu.suda.edu.cn \\
    \{wangzhefeng, duanxinyu\}@huawei.com,
    zhli13@suda.edu.cn\\
}

\abstract{
This work revisits the topic of jointly parsing constituency and dependency trees, i.e., to produce compatible constituency and dependency trees simultaneously for input sentences, which is attractive considering that the two types of trees are complementary in representing syntax. 
The original work of \citet{zhou-zhao-2019-head} performs joint parsing only at the inference phase. They train two separate parsers under the multi-task learning framework (i.e., one shared encoder and two independent decoders).  
They design an ad-hoc dynamic programming-based decoding algorithm of $O(n^5)$ time complexity for finding optimal compatible tree pairs. 
Compared to their work, we make progress in three aspects: (1) adopting a much more efficient decoding algorithm of $O(n^4)$ time complexity, (2) exploring joint modeling at the training phase, instead of only at the inference phase, (3) proposing high-order scoring components to promote constituent-dependency interaction. We conduct experiments and analysis on seven languages, covering both rich-resource and low-resource scenarios. Results and analysis show that joint modeling leads to a modest overall performance boost over separate modeling, but substantially improves the complete matching ratio of whole trees, thanks to the explicit modeling of tree compatibility. 
 \\ \newline  
}

\begin{document}

\maketitleabstract

\section{Introduction}
\label{sec:intro}

\begin{figure}[tb!]
  \centering
  \small
    \subfigure[Constituency tree]{
        \label{fig:cosnt}    
        \begin{tikzpicture}[
              level distance=22.5pt,
              every tree node/.style={align=center,anchor=base},
              frontier/.style={distance from root=70pt},
              case_red/.style={brickred, fill=brickred!20, draw=brickred, thick, fill opacity=1., text opacity=1., rounded corners=1mm,align=center, minimum size=4mm},
              case_blue/.style={midnightblue, fill=midnightblue!20, draw=midnightblue, thick, fill opacity=1., text opacity=1., rounded corners=1mm,align=center, minimum size=4mm},
              edge from parent/.style={thick, draw, black!70, edge from parent path={(\tikzparentnode.south) {[rounded corners=0.5pt]-- ($(\tikzchildnode |- \tikzparentnode.south) + (0, -5pt)$) -- (\tikzchildnode)}}} 
              ]
        \Tree
            [.S 
                [.NP Logic$_1$ ]
                [.VP 
                    plays$_2$ 
                    [.NP
                        a$_3$ 
                        maximal$_4$ 
                        role$_5$ 
                    ]
                    [.ADVP here$_6$ ]
                ]
            ];
        \end{tikzpicture}
    }
    \hfill
        \subfigure[Dependency tree]{
        \label{fig:dep}
        \begin{dependency}[arc angle=80]
            \begin{deptext} [row sep=0.2cm, column sep=0.1cm]
            \$ \& Logic$_1$ \& plays$_2$ \& a$_3$ \& maximal$_4$ \& role$_5$ \& here$_6$ \\
            \end{deptext}

            \depedge[edge style={black!70, thick}, edge height=2ex]{3}{2}{nsubj}
            \depedge[edge style={black!70, thick}, edge height=5ex]{1}{3}{root}   
            \depedge[edge start x offset=2pt, edge style={black!70, thick}, edge height=3ex]{6}{4}{det}
            \depedge[edge style={black!70, thick}, edge height=1ex]{6}{5}{amod}
            \depedge[edge start x offset=1pt, edge end x offset=1pt, edge style={black!70, thick}, edge height=6ex]{3}{6}{dobj}
            \depedge[edge start x offset=-1pt, edge style={black!70, thick}, edge height=9ex]{3}{7}{advmod}

        \end{dependency}
    }
  \caption{Constituency and dependency trees. 
  }
  \label{fig:con_a_dep}
\end{figure}
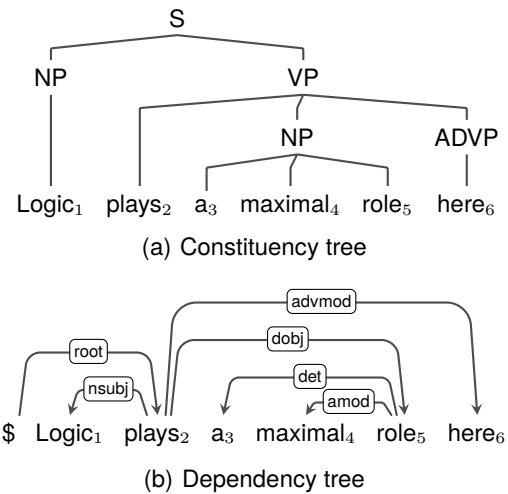

As one of the most fundamental and long-standing NLP tasks, syntactic parsing aims to reveal how sentences are syntactically structured. Among many paradigms for representing syntax, constituency trees (c-trees) and dependency trees (d-trees) are the most popular and have gained tremendous research attention in both data annotation and parsing techniques. Figure \ref{fig:con_a_dep} gives example trees. 

As is well known, c-trees and d-trees capture syntactic structure from different yet complementary perspectives \cite{book_treebanking}. 
On the one hand, c-trees can clearly illustrate how sentences are composed hierarchically. Constituents, especially major phrases like NPs and VPs, are often self-evident and thus easily agreed upon by people. 
On the other hand, d-trees emphasize pairwise syntactic (sometimes even semantic) relationships between words, i.e., what role (function) the modifier word plays for the head word. It is usually more simple and flexible to draw 
dependency links than to add constituent nodes. 

Therefore, it can be an attractive and useful feature that a parsing model outputs both c-trees and d-trees at the same time. %
Of course the two trees must be compatible with each other. Basically, \textbf{compatibility} means that for any constituent, only the single head word can compose dependency links, either inwards or outwards, with words outside the constituent. 
Please note that this work and most previous works consider only conventional continuous c-trees and projective d-trees.\footnote{In the sentence ``A hearing is scheduled tomorrow on this issue'', ``A hearing on this issue''  composes a typical discontinuous constituent, which also results in a non-projective dependency tree. }

From the modeling perspective, parsing techniques have achieved immense progress, thanks to the development of deep learning techniques, especially of the pre-trained language models (PLMs) \cite{peters-etal-2018-deep, devlin-etal-2019-bert}.  
One amazing characteristic is that constituency parsing and dependency parsing have a nearly identical model architecture nowadays. Under the graph-based parsing framework, besides the encoder, the scoring components are the same as well if we treat a constituent as a link between the beginning word and the end word \cite{zhang-etal-2020-fast}. 
The only difference lies in the decoding algorithms for finding optimal c- or d-trees. 
To some extent, such development stimulates research on joint constituency and dependency parsing \citep{zhou-zhao-2019-head}.

The original work of \citet{zhou-zhao-2019-head} performs joint parsing \emph{only at the inference phase}. First, they use the simplified head-driven phrase structure grammar (HPSG) trees to encode both c- and d-trees simultaneously. In fact, their HPSG trees are inherently the same as lexicalized constituency trees (l-trees), in which each constituent is annotated with its head word \citep{collins-2003-head}. 
Figure \ref{fig:con_a_lex} gives an example l-tree. 
Then, they design an ad-hoc decoding algorithm of $O(n^5)$ time complexity for finding optimal l-trees. 
Under the multi-task learning (MTL) framework, they use one shared encoder, and two separate decoders for the two parsing tasks respectively. 
Thus there is no explicit interaction between the two types of parse trees.

In this paper, we carry forward this interesting research topic from the following three aspects. 
\begin{enumerate}
    \item Inspired by the recent work of \citet{lou-etal-2022-nested} on nested named entity recognition (NER), we employ the  Eisner-Satta algorithm \citep{eisner-satta-1999-efficient} of $O(n^4)$ time complexity for finding optimal l-trees, which runs very efficiently on GPUs after proper batchification \cite{zhang-etal-2020-efficient}. Experiments show that our model is about 2.5$\times$ faster than that of \citet{zhou-zhao-2019-head}.
    \item We propose to jointly model the two parsing tasks at both training and inference phases. 
    \item We propose high-order scoring components so that the two types of parse trees can interact with and influence each other more tightly. 
\end{enumerate} 
We conduct experiments on benchmark datasets in seven languages, covering both rich-resource and low-resource scenarios. The results and analysis lead to several interesting findings. We release our code at \href{https://github.com/EganGu/high-order-joint-parsing}{https://github.com/EganGu/high-order-joint-parsing}.

\section{Related Work}

\textbf{Before the deep learning era}, there exist three competitive families of constituency parsing approaches, i.e., lexicalized PCFG parsing \cite{collins-1999-head}, unlexicalized PCFG parsing \cite{petrov-etal-2006-learning}, and discriminative shift-reduce parsing \cite{zhu-etal-2013-fast}. Due to the importance of head word features, all three families are more or less related to our work, in the sense that the parser may output d-trees as byproduct at the inference phase.

Lexicalized PCFG parsers generate a c-tree in a head-centered, top-down manner \cite{collins-1999-head}. The head token of each constituent is settled when parsing is finished. Therefore, it is straightforward to acquire \emph{unlabeled} d-trees. 

The unlexicalized parser of \citet{klein-manning-2003-accurate} splits each constituent labels into multiple sub-labels according to linguistic heuristics. 
For N-ary production rules, they markovize out from the head child, making it feasible to recover unlabeled d-trees likewise. 

The discriminative shift-reduce parser of \citet{crabbe-2015-multilingual} makes heavy use of head tokens for composing features, and therefore is also capable of producing unlabeled d-trees.  

Our work is also closely related to \citet{klein-2002-fast}, who factorize a l-tree into an unlexicalized c-tree and a d-tree, corresponding to two generative models that are separately trained. 
They propose an efficient yet inexact A* algorithm for finding the optimal l-tree, instead of using the Eisner-Satta algorithm.

\textbf{In the deep learning era}, besides \citet{zhou-zhao-2019-head}, there exist two works that tackle constituency parsing and dependency parsing simultaneously. 

\citet{strzyz-etal-2019-sequence} transform both constituency parsing and dependency parsing into sequence labeling tasks. However, the parsing performance lags behind state-of-the-art models by large margins. 
\citet{fer-2022} transform constituency parsing into a dependency parsing task and employ a pointer network architecture to obtain the dependency trees. 

The above two works have two key differences from our work. 
First, during training, 
both works employ the MTL framework, hence blocking explicit interaction between two sub-modules. 
Second and more importantly, both works do not consider the compatibility of output c-trees and d-trees at the inference phase.

\section{Lexicalized Tree Representation}
\label{sec:l-tree-repr}

Given an input sentence $\boldsymbol{x} = w_1 \dots w_n$, we use $\{(i, j, l), 0 \leq i, j \le n\}$ to denote a c-tree $\boldsymbol{c}$,  and $\{(h \rightarrow m, r), 0 \leq h, m \le n\}$ to a d-tree $\boldsymbol{d}$. 
For a c-tree, $(i,j,l)$ denotes a constituent spanning $w_{i} \dots w_{j}$ and labeled as $l \in \mathcal{L}$, while
for a d-tree, $(h \rightarrow m,r)$ denotes a dependency from the head word $w_h$ to the modifier word $w_m$ and labeled as $r \in \mathcal{R}$.

\begin{figure}[tb]
  \centering
  \small
  \pgfkeys{/pgf/inner sep=0.1em}
    \subfigure[A regular c-tree (head-binarized)]{
        \label{fig:con-cnf}    
        \begin{tikzpicture}[
              scale=0.9,
              level distance=18pt,
              every tree node/.style={align=center,anchor=base},
              frontier/.style={distance from root=105pt},
              case_red/.style={brickred, fill=brickred!20, draw=brickred, thick, fill opacity=1., text opacity=1., rounded corners=1mm,align=center, minimum size=4mm},
              case_blue/.style={midnightblue, fill=midnightblue!20, draw=midnightblue, thick, fill opacity=1., text opacity=1., rounded corners=1mm,align=center, minimum size=4mm},
              edge from parent/.style={thick, draw, black!70, edge from parent path={(\tikzparentnode.south) {[rounded corners=0.5pt]-- ($(\tikzchildnode |- \tikzparentnode.south) + (0, -5pt)$) -- (\tikzchildnode)}}} 
              ]
        \Tree
            [.S 
                [.NP Logic$_1$ ]
                [.VP 
                    [.VP$^\ast$ 
                        [.VP$^\ast$ plays$_2$ ]
                        [.NP
                            [.NP$^\ast$ a$_3$ ] 
                            [.NP$^\ast$ 
                                [.NP$^\ast$ maximal$_4$ ] 
                                [.NP$^\ast$ role$_5$ ]
                            ]
                        ]
                    ]
                    [.ADVP here$_6$ ]
                ]
            ];
        \end{tikzpicture}
    }
    \hfill
    \subfigure[A lexicalized c-tree]{
        \label{fig:lex}    
        \begin{tikzpicture}[
              scale=0.9,
              level distance=18pt,
              every tree node/.style={align=center,anchor=base},
              frontier/.style={distance from root=105pt},
              case_red/.style={brickred, fill=brickred!20, draw=brickred, thick, fill opacity=1., text opacity=1., rounded corners=1mm, align=center, minimum size=4mm},
              case_blue/.style={midnightblue, fill=midnightblue!20, draw=midnightblue, thick, fill opacity=1., text opacity=1., rounded corners=1mm, align=center, minimum size=4mm},
              edge from parent/.style={thick, draw, black!70, edge from parent path={(\tikzparentnode.south) {[rounded corners=0.5pt]-- ($(\tikzchildnode |- \tikzparentnode.south) + (0, -5pt)$) -- (\tikzchildnode)}}} 
              ]
        \Tree
            [.S[plays] 
                [.NP Logic$_1$ ]
                [.VP[plays] 
                    [.VP$^\ast$[plays] 
                        [.VP$^\ast$ plays$_2$ ]
                        [.NP[role]
                            [.NP$^\ast$ a$_3$ ] 
                            [.NP$^\ast$[role] 
                                [.NP$^\ast$ maximal$_4$ ] 
                                [.NP$^\ast$ role$_5$ ]
                            ]
                        ]
                    ]
                    [.ADVP here$_6$ ]
                ]
            ];
        \end{tikzpicture}
    }
  \caption{Example %
  c-trees. The head word of each lexicalized constituent is marked by $\left[ \cdot \right]$.}
  \label{fig:con_a_lex}
\end{figure}
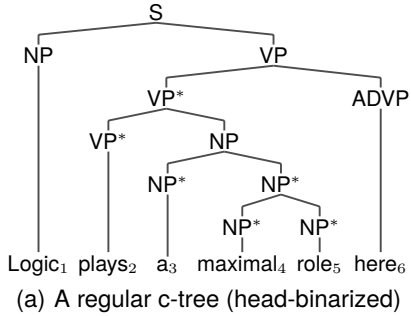
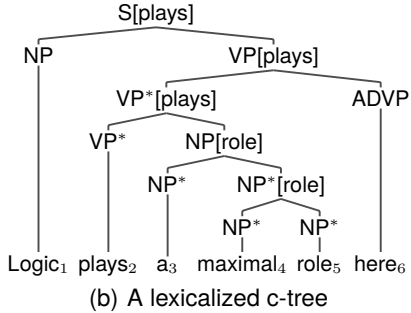

It is a natural way to 
encode both a c-tree $\boldsymbol{c}$ and a d-tree $\boldsymbol{d}$ into an l-tree as the joint representation \citep{collins-2003-head}. 
Figure \ref{fig:con_a_lex} gives examples. 
We convert the original c-tree into Chomsky normal form (CNF), as shown in Figure \ref{fig:con-cnf}, which is required by the graph-based parsing model adopted in this work. 
Figure \ref{fig:lex} presents the corresponding l-tree, in which the head words are decided by the d-tree in Figure \ref{fig:dep}. 

Formally, we denote an l-tree as $\boldsymbol{t} = \{(i, j, h, l)\}$, where $(i, j, h, l)$ represents a lexicalized constituent spanning $w_{i} \dots w_j$ with a head word $w_h (i\leq h\leq j)$ and a label $l \in \mathcal{L}$. 
We use $\mathrm{NP[role]_{3, 5}}$ as a simplified notation of $(3, 5, 5, \mathrm{NP})$.

\paragraph{Constituent-to-dependency conversion and the compatibility rationale. } 
To the best of our knowledge, many dependency treebanks are automatically converted from constituent treebanks, instead of constructed from scratch. 
Usually, the const-to-dep conversion process consists of two major steps: 1) determining unlabeled dependencies using head-finding rules, and 2) determining dependency labels \citep{de-marneffe-etal-2006-generating}. In the first step, the head-finding rules are applied on c-trees to form l-trees, in which each constituent has only one head word. Further, unlabeled d-trees are constructed according to head words.  
Such conversion process determines that the resulting dependency trees are compatible with the original constituent trees. 

\paragraph{Head-binarization.} As shown earlier, given a c-tree $\boldsymbol{c}$ and a d-tree $\boldsymbol{d}$, we first convert $\boldsymbol{c}$ into CNF before  constructing an l-tree $\boldsymbol{t}$, which is required by the CKY algorithm. 

To avoid incompatibility after CNF conversion, we adopt the head-binarization strategy \citep{collins-2003-head}, instead of left- or right-binarization. Basically, head-binarization dynamically selects left- or right-binarization to comply with a given d-tree.
For example, the constituent $\mathrm{VP[plays]}_{2, 6}$ in Figure \ref{fig:lex} is left-binarized, whereas $\mathrm{NP[role]}_{3, 5}$ is right-binarized. If both binarization choices are feasible, we prioritize left-binarization.  

\paragraph{Recovering constituency/dependency trees.}
It is obvious that an l-tree becomes a c-tree when discarding the head words.
Meanwhile, since l-trees are strictly binarized, we can construct the corresponding d-trees according to the head words. 

Moreover, please note that dependency relations are not encoded in an l-tree $\boldsymbol{t}$, and are handled via a separate labeling step (see Section \ref{sec:two-stage-parsing}), which is now common practice in parsing models \cite{zhang-etal-2020-efficient}. 

\paragraph{Labeling compatibility. } Since the dependency relations are not contained in l-trees, our joint representation can only guarantee that the encoded c- and d-trees are compatible in an unlabeled fashion. 
Given that the dependency relations are usually derived from c-trees, it leads to an interesting question: \textit{is it possible to make the parsed c- and d-trees compatible at the labeled level according to the const-to-dep conversion process?} We believe the answer is no. We give our explanations as follows. 

During the two steps of the conversion process, the label determination algorithm makes use of non-local constituent features (i.e., multiple constituents and their labels). Therefore, it is actually difficult to impose such labeling compatibility into our dynamic programming decoding (described in Section \ref{sec: inference}). 
Moreover, some const-to-dep conversion processes make use of not only structural constituent labels like NP/VP, but also functional labels like -SBJ/OBJ \citep{surdeanu-etal-2008-conll, seddah-etal-2014-introducing}. Functional labels are usually overlooked in constituency parsing research.

\section{The Joint Parsing Approach}

In this section, we first describe a first-order model, and then introduce a second-order extension. 

To facilitate explanation, we give alternative notations. We use $\boldsymbol{y}=\{(i, j, h)\}$ to denote an unlabeled l-tree, i.e., $\boldsymbol{t}$ without the constituent labels, $\boldsymbol{l}$ to denote the set of constituent labels in an l-tree $\boldsymbol{t}$, and $\boldsymbol{r}$ to denote the set of dependency relations in a d-tree $\boldsymbol{d}$. 
Then we have the following equivalent notations. 
$$(\boldsymbol{c}, \boldsymbol{d}) \equiv (\boldsymbol{t}, \boldsymbol{r}) \equiv (\boldsymbol{y}, \boldsymbol{l}, \boldsymbol{r})$$
Please kindly note that we assume that $\boldsymbol{c}$ is binarized and conforms to CNF, which guarantees that a unique unlabeled dependency tree $\boldsymbol{d}$ can be induced from $\boldsymbol{t}$ and $\boldsymbol{y}$.

\subsection{Two-stage Parsing and First-order Factorization}\label{sec:two-stage-parsing}

Following \citet{dozat-2017-deep} and \citet{zhang-etal-2020-fast}, we adopt the two-stage parsing strategy, which has been proven to be able to simplify the model architecture and improve efficiency, without hurting performance.

\paragraph{Stage \uppercase\expandafter{\romannumeral1}: Lexicalized Bracketing.} Given $\boldsymbol{x}$, the goal of the first stage is to find an optimal unlabeled l-tree $\boldsymbol{y}$, whose score is:
\begin{equation}
s(\boldsymbol{x}, \boldsymbol{y}) =  \sum_{(i,j) \in \boldsymbol{c}} s^c (i,j)
+ \sum_{h \rightarrow m \in \boldsymbol{d}} s^d (h,m)
\label{eq.score}
\end{equation}
where $s^{c/d}$ denote the scores of unlabeled constituents and unlabeled dependencies, respectively. Here we adopt the first-order factorization, i.e., the scores of constituents and dependencies being mutually independent. We present second-order factorization in Section \ref{sec:second-order}.

Given all $s^c(i,j)$ and $s^d(h,m)$, the decoding algorithm determines the 1-best l-tree. 
\begin{equation}
    \hat{\boldsymbol{y}} = \mathop{\arg \max}\limits_{\boldsymbol{y} \in \mathcal{Y}(\boldsymbol{x})} {s}(\boldsymbol{x}, \boldsymbol{y})
\end{equation}
where $\mathcal{Y}(\boldsymbol{x})$ denotes the set of all legal l-trees for $\boldsymbol{x}$.

\paragraph{Stage \uppercase\expandafter{\romannumeral2}: Labeling.} Given unlabeled constituents/dependencies in $\hat{\boldsymbol{y}}$ obtained in the first stage, this stage independently predicts labels for them. 
\begin{align} \label{equation:label-argmax}
&\hat{l} = \mathop{\arg \max}\limits_{%
l \in \mathcal{L}} s^c(i,j,l) \\
&\hat{r} = \mathop{\arg \max}\limits_{%
r \in \mathcal{R}} s^d(h,m,r)
\end{align}

\subsection{Training Loss}
Given a training instance $(\boldsymbol{x}, \boldsymbol{y}, \boldsymbol{l}, \boldsymbol{r})$, the training loss consists of two parts, corresponding to the two stages.\footnote{
At a training step, 
the cumulative loss of all instances in a mini-batch is divided by the total number of tokens.
}
\begin{equation}
L(\boldsymbol{x}, \boldsymbol{y}, \boldsymbol{l}, \boldsymbol{r})=L^{\mathrm{bracket}}(\boldsymbol{x}, \boldsymbol{y})+L^{\mathrm{label}}(\boldsymbol{x}, \boldsymbol{y}, \boldsymbol{l}, \boldsymbol{r})
\end{equation}

For the first stage, we adopt the max-margin loss \citep{taskar-etal-2004-max}\footnote{We also try using the CRF loss for the l-tree and find it to be slightly inferior to the max-margin loss in terms of performance and training speed.}, which is based on the margin between the scores of the gold-standard l-tree $\boldsymbol{y}$ and the predicted one $\hat{\boldsymbol{y}}$. 

\begin{equation}
\begin{split}
&L^{\mathrm{bracket}} = \\
&\quad %
\max \left( 0, \mathop{\max}\limits_{\boldsymbol{y} \neq \hat{\boldsymbol{y}}} \left( s(\boldsymbol{y}) - s(\hat{\boldsymbol{y}}) + \Delta(\boldsymbol{y}, \hat{\boldsymbol{y}}) \right)\right)
\end{split}
\label{eq.mm}
\end{equation}
where $\Delta$ is the Hamming distance, which we set as the number of incorrect dependencies and constituents.

 \vspace{6pt} 
For the second-stage loss,  following \citet{dozat-2017-deep}, we employ local cross-entropy losses, i.e., the sum of classification loss for labeling all constituents and dependencies in the gold- standard c-tree and d-tree.

Besides, %
we tried to balance the losses from the two stages via weighted summation, which we found has negligible influence on performance according to our preliminary experiments. 

\subsection{Inference: \textcolor{black}{Batchified Eisner-Satta}}
\label{sec: inference}
\citet{zhou-zhao-2019-head} propose a naive CKY-style algorithm in $O(n^5)$ complexity to the simplified HPSG inference, which is highly time-consuming. For l-trees, \citet{eisner-satta-1999-efficient} propose a relatively fast inference that reduces the complexity to $O(n^4)$ by merging dependency arcs in advance. In this paper, we adopt the Eisner-Satta algorithm to jointly infer the c- and d-trees. The deduction rules \citep{pereira-warren-1983-parsing} of the algorithm are illustrated in Figure \ref{fig:satta}.

To further increase the speed of inference, following \citet{zhang-etal-2020-efficient}, we batchify the Eisner-Satta algorithm to fully utilize GPUs, as shown in Algorithm \ref{alg:decode}. 
The basic idea is handling spans of the same length simultaneously. 
The time complexity of our algorithm is practically linear.

\begin{algorithm*}[tb]
\caption{The Eisner-Satta Decoding Algorithm.}
\begin{algorithmic}[1]
\newlength{\commentindent}
\setlength{\commentindent}{.42\textwidth}
\renewcommand{\algorithmiccomment}[1]{\unskip\hfill\makebox[\commentindent][l]{$\rhd$~#1}\par}
\renewcommand{\algorithmic}[1][0]{%
  \oldalgorithmic[#1]%
  \renewcommand{\ALC@com}[1]{%
  \ifnum\pdfstrcmp{##1}{default}=0\else\algorithmiccomment{##1}\fi}%
}

\begin{footnotesize}
\STATE \textbf{define:} $\alpha, \beta \in \mathbb{R}^{n \times n \times n \times B}$ \COMMENT{$B$ is \#sents in a batch}
\STATE \textbf{initialize:} all $\alpha^{:, :, :} = 0$, $\beta^{:, :, :} = 0$\\
\FOR [span width]{$w = 1$ \TO $n$}
    \STATE \emph{{\rm Batchify}: $0 \le i,j \le n,j=i+w, k,h$}
    \STATE $\alpha^{i, j, :} = s^c(i, j)+\mathop{\max}\limits_{i \le k < j}\left(\alpha^{i, k, :}+\beta^{k+1, j, :}, \beta^{i, k, :}+\alpha^{k+1, j, :} \right)$ \COMMENT{add $s(i, j,:)$ to $\alpha/\beta$ for the 2rd-order extension}
    \STATE $\beta^{i, j, :} = \mathop{\max}\limits_{i \le h \le j}\left(\alpha^{i, j, h}+s^d(:, h)\right)$

\ENDFOR
\RETURN $\alpha^{0, n-1, 0}$ %
\end{footnotesize}
\end{algorithmic}
\label{alg:decode}
\end{algorithm*}

\subsection{Second-order Extension} \label{sec:second-order}

\begin{figure}
    \centering
    \includegraphics[scale=0.22]{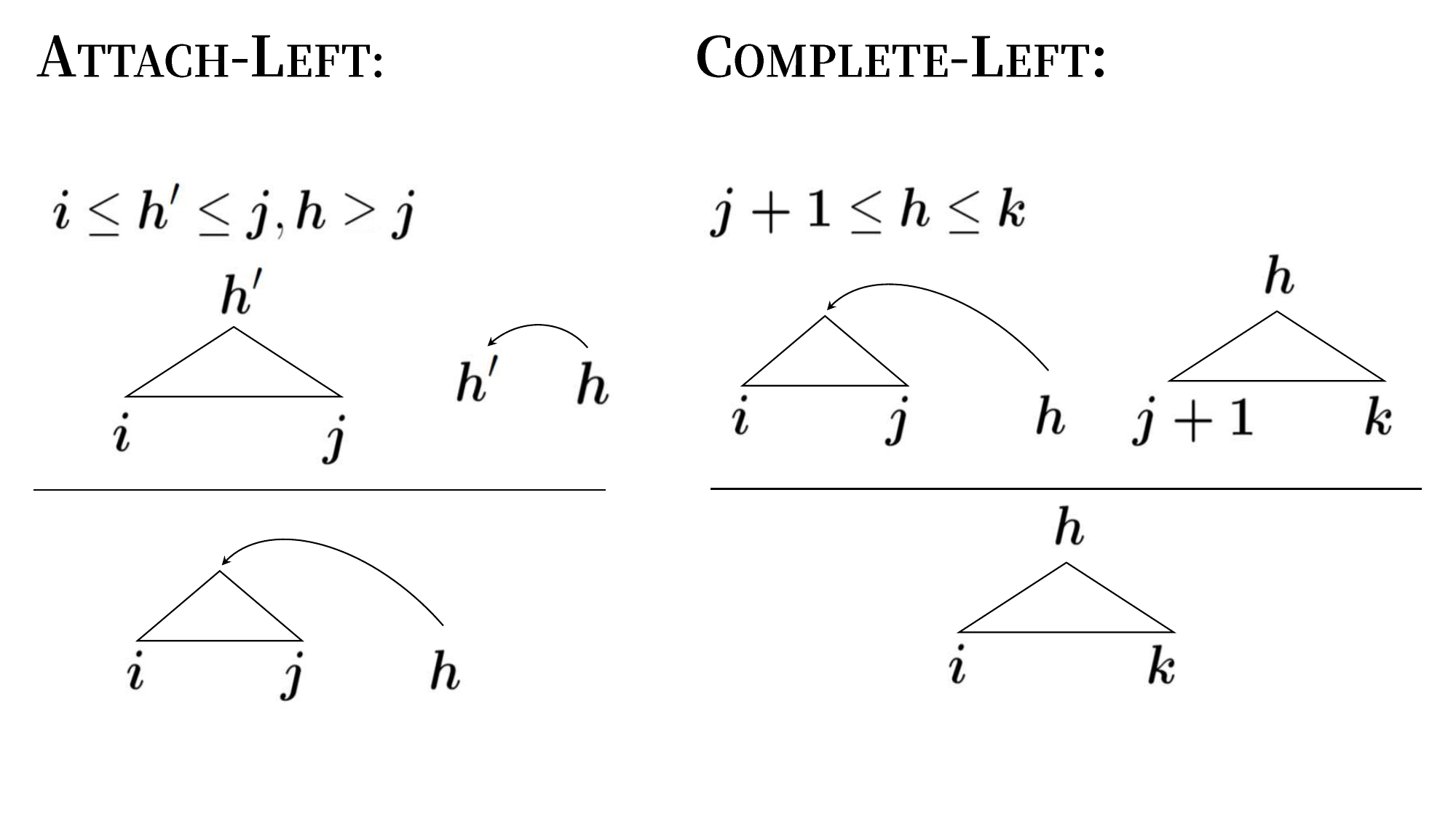}
    \caption{Deduction rules for Eisner-Satta algorithm \citep{eisner-satta-1999-efficient}. 
    We show only the leftward rules, omitting the symmetric rightward one as well as initial conditions for brevity.}
    \label{fig:satta}
\end{figure}

In order to further bridge the c- and d-trees for more in-depth joint modeling, we extend the score decomposition of $\boldsymbol{y}$ to the second-order case. 
We define two types of span structure, as 
\begin{itemize}
    \item Headed spans: a span $(i, j)$ with $h$ being the lexical head (parent, or root), in which $i \le h \le j$. 
    \item Hooked spans: a span $(i, j)$ with $h$ being the lexical grandparent, in which $h < i ~\textit{or}~ h > j$. 
\end{itemize}
Both headed spans and hooded spans can be denoted as $(i, j, h)$, and can be distinguished by the position of $h$. 

As shown in Figure \ref{fig:satta}, the left part illustrates the construction of a hooked span, whereas the right part illustrates how to create a headed span.

We further incorporate scores of headed spans and hooked spans into the basic first-order model. 
\begin{equation}
s^{2o}(\boldsymbol{y}) = s(\boldsymbol{y}) + \sum_{(i,j, h) \in \boldsymbol{y}}s^{\mathrm{span}}(i,j, h)
\label{eq.sec-score}
\end{equation}
where $s(i,j, h) \in \mathbb{R}^{n \times n \times n}$.

\section{Model Architecture}

In this section, we introduce how to compute scores of lexicalized subtrees.  

\paragraph{Encoding.}
Given a sentence $\boldsymbol{x} = w_0, \dots, w_{n+1}$, where $w_0$ and $w_{n+1}$ denote the \textit{<bos>} (begin of a sentence) and \textit{<eos>} (end of a sentence) tokens respectively, 
the corresponding hidden representations $\boldsymbol{e}$ can be obtained via BERT encoder \citep{devlin-etal-2019-bert}, without cascading word embedding and LSTM layers for simplicity.

\paragraph{Scoring.} We employ the same scoring functions as \citet{zhang-etal-2020-efficient, zhang-etal-2020-fast} to compute span scores $s^c(i,j)$ and arc scores $s^d(h,m)$. 
\begin{align}
&r^{\mathrm{left/right}}_i = \mathrm{MLP}^{\mathrm{left/right}}(\overrightarrow{e_i} \oplus \overleftarrow{e_{i+1}}) \\
&r^{\mathrm{head/mod}}_i = \mathrm{MLP}^{\mathrm{head/mod}}(\overrightarrow{e_i} \oplus \overleftarrow{e_i}) \\ 
&s^c(i,j) =  
\begin{bmatrix}
r_{i}^{\mathrm{left}} \oplus 1 
\end{bmatrix}^\mathrm{T}
\boldsymbol{W}^c r_{j}^{\mathrm{right}} \\
&s^d(h,m) =  
\begin{bmatrix}
r_{m}^{\mathrm{mod}} \oplus 1 
\end{bmatrix}^\mathrm{T}
\boldsymbol{W}^d r_{h}^{\mathrm{head}}
\end{align}
where MLPs are applied to obtain lower $k$-dimensional vectors; $r^{\mathrm{head/mod}}_i$ are the representation vectors of $w_i$ as a head/modifier  respectively, and analogously, $r^{\mathrm{left/right}}_i$ are the left/right boundary representation of $w_i$; $\overrightarrow{e_i}/\overleftarrow{e_i}$ are hidden representations of $w_i$ in different contextual orientations\footnote{For BERT encoder, we split each $e_i \in \boldsymbol{e}$ in half to construct each $(\overrightarrow{e_i}, \overleftarrow{e_i})$ the same as \citet{kitaev-klein-2018-constituency}.
}. $\mathbf{W}^{d/c} \in \mathbb{R}^{(k+1) \times k}$ are trainable parameters.

For second-order extension, we follow  \citet{yang-tu-2022-headed} to calculate scores of headed spans and hooked spans. 
\begin{align}
&r^{\mathrm{word}}_i = \mathrm{MLP}^{\mathrm{word}}(\overrightarrow{e_i} \oplus \overleftarrow{e_i}) \\ 
&h^{\mathrm{span}}_{i, j} = 
\left[\overrightarrow{e_i} \oplus \overleftarrow{e_{i+1}}\right] - \left[\overrightarrow{e_j} \oplus \overleftarrow{e_{j+1}}\right]\\
&r^{\mathrm{span}}_{i, j} = \mathrm{MLP}^{\mathrm{span}}(h^{\mathrm{span}}_{i, j}) \\ 
&s^\mathrm{span}(i,j,h) =  
\begin{bmatrix}
r_{h}^{\mathrm{word}} \oplus 1 
\end{bmatrix}^\mathrm{T}
\boldsymbol{W}^{\mathrm{span}}
\begin{bmatrix}
r_{i, j}^{\mathrm{span}} \oplus 1 
\end{bmatrix}
\end{align}
where $\boldsymbol{W}^{\mathrm{span}} \in \mathbb{R}^{(k+1) \times (k+1)}$. %
Please notice that the computation of scores of headed spans and hooked spans share the same parameters. Our preliminary experiments show that the performance changes little if we use separate parameters. This may indicate that the model can distinguish the two types of scores only according to the position of $h$.

For the second-stage labeling tasks, we follow \citet{zhang-etal-2020-efficient, zhang-etal-2020-fast} to  
calculate dependency and constituency labels scores.

\section{Experiments}

\subsection{Settings} 
\paragraph{Datasets.}
We conduct experiments on the English Penn Treebank (PTB) \citeplanguageresource{ptb},
the Chinese Penn Treebank 5.1 (CTB) \citeplanguageresource{ctb} and the SPMRL datasets \citep{seddah-etal-2014-introducing}.

For PTB, we follow the standard splits, i.e., sections 02-21 for training, section 22 for development and section 23 for testing. The d-trees are obtained with Stanford Typed Dependencies (SD) \citep{de-marneffe-etal-2006-generating} using the Stanford parser v3.3.0\footnote{https://nlp.stanford.edu/software/lex-parser.shtml}.

For CTB, we use the same split as \citet{zhang-clark-2008-tale}. The c-trees are converted to the corresponding d-trees using the Penn2Malt tool\footnote{https://cl.lingfil.uu.se/\textasciitilde nivre/research/Penn2Malt.html}. 

For SPMRL, we adhere to the default data split and utilize the provided parallel constituency and dependency treebanks. 
We specifically choose five languages from SPMRL—French, Hebrew, Korean, Polish, and Swedish—where d-trees are well compatible with c-trees.
Based on different characteristics of languages, the multilingual d-trees are obtained from various const-to-dep conversion processes.

Table \ref{table:statistics} presents the data scale and compatibility of different languages. The languages are categorized into two scenarios: rich- and low-resource. We notice that in almost all incompatible cases, multiple words within the same constituent are headed by words outside the
constituent, leading to the failure of building valid l-trees. 
To address this issue,  we remove all incompatible instances from the training sets.

\begin{table}[tb]
\footnotesize
\setlength{\tabcolsep}{2.pt}
\centering
\begin{tabular*}
    {\columnwidth}{@{\extracolsep{\fill}}l lll}
    \toprule
    & Train & Dev & Test \\
    \midrule
    \multicolumn{4} {c}{\emph{Rich resource}} \\ 
    {\bf English}    &  39.8K (99.8)  &  1.7K (99.7)    &  2.4K (99.9) \\
    {\bf Chinese}    &  16.1K (99.9)  &  0.8K (99.8)    &  1.9K (99.8) \\
    {\bf French}     &  14.7K (99.5)  &  1.2K (99.6)    &  2.5K (99.7) \\
    {\bf Korean}     &  23.0K (100.0) &  2.1K (100.0)   &  2.3K (100.0) \\
    \multicolumn{4} {c}{\emph{Low resource}} \\ 
    {\bf Hebrew}     &  5.0K (94.9)   &  0.5K (95.0)    &  0.7K (96.9) \\
    {\bf Polish}     &  6.6K (95.1)   &  0.8K (94.1)    &  0.8K (94.2)     \\
    {\bf Swedish}    &  5.0K (92.0)   &  0.5K (88.6)    &  0.7K (94.1)     \\
    \bottomrule
    \end{tabular*}
    \caption{Number of sentences and percentage of compatible sentences (in parentheses).}
    \label{table:statistics}
\end{table}

\begin{table*}[tb!]
    \centering 
    \footnotesize
    \setlength{\tabcolsep}{0.28em}
    \begin{tabular*}
        {\textwidth}{l cc cc cc cc cc cc cc cc}
        \toprule
        & \multicolumn{2}{c}{\footnotesize {\bf English}} & \multicolumn{2}{c}{\footnotesize {\bf Chinese}} & \multicolumn{2}{c}{\footnotesize {\bf French}} & \multicolumn{2}{c}{\footnotesize {\bf Hebrew}} & \multicolumn{2}{c}{\footnotesize {\bf Korean}} & \multicolumn{2}{c}{\footnotesize {\bf Polish}} & \multicolumn{2}{c}{\footnotesize {\bf Swedish}} & \multicolumn{2}{c}{\footnotesize {\it Average}} \\
        \cmidrule(lr){2-3}\cmidrule(lr){4-5}\cmidrule(lr){6-7}\cmidrule(lr){8-9}\cmidrule(lr){10-11}\cmidrule(lr){12-13}\cmidrule(lr){14-15}\cmidrule(lr){16-17}
        & LAS & F1 & LAS & F1& LAS & F1& LAS & F1& LAS & F1& LAS & F1& LAS & F1& LAS & F1 \\
        \midrule 
       SEP &$ 95.55 $&$ 95.92 $&$ 90.55 $&$ 90.79 $&$ 89.31 $&$ 87.51 $&$ 85.12 $&$ \pmb{93.15} $&$ 89.67 $&$ 89.44 $&$ \pmb{91.20} $&$ 96.21 $&$ 87.94 $&$ 89.70 $&$ 89.91 $&$ 91.82$  \\ 

       MTL &$ 95.59 $&$ \pmb{95.97} $&$ 90.86 $&$ \pmb{90.83} $&$ 89.42 $&$ \pmb{87.58} $&$ 85.77 $&$ 93.09 $&$ \pmb{89.93} $&$ \pmb{89.46} $&$ 90.83 $&$ 96.33 $&$ 88.11 $&$ \pmb{89.73} $&$ 90.07 $&$ \pmb{91.86}$ \\ 
       
       Joint1o &$ 95.62 $&$ 95.76 $&$ 90.88 $&$ 90.77 $&$ 89.50 $&$ 87.25 $&$ 85.63 $&$ 92.93 $&$ 89.83 $&$ 89.18 $&$ 90.80 $&$ 96.29 $&$ 88.32 $&$ 89.52 $&$ 90.08 $&$ 91.67 $\\ 
       
       Joint2o &$ \pmb{95.64} $&$ 95.80 $&$ \pmb{90.96} $&$ 90.76 $&$ \pmb{89.52} $&$ 87.48 $&$ \pmb{85.83} $&$ 92.96 $&$ 89.83 $&$ 89.21 $&$ 91.15 $&$ \pmb{96.37} $&$ \pmb{88.38} $&$ 89.69 $&$ \pmb{90.19} $&$ 91.75$ \\ 
       \bottomrule 
    \end{tabular*}
    \caption{Results on multilingual test sets (including the PTB, CTB and SPMRL). }
    \label{exp:cmp_ml}
\end{table*}

\paragraph{Evaluation metrics.} For dependency parsing, we adopt unlabeled and labeled attachment scores (UAS/LAS) as metrics. For constituency parsing, we adopt the standard constituent-level labeled precision (P), recall (R), and F1-score as metrics. 
Consistent with \citet{zhou-zhao-2019-head}, we omit all punctuation for dependency parsing. 

\paragraph{Models.} We conduct experiments with our joint parsing model: \textbf{Joint1o} and \textbf{Joint2o}, which both perform lexicalized modeling but with the first/second-order scores separately. 
To facilitate a comprehensive comparison, we introduce two kinds of baseline models, using the same network architectures as joint models: 
\begin{itemize}
    \item \textbf{Separate parsing models (SEP)}. We train two separate constituency and dependency parsers. Following joint models, we also apply max-margin loss for each task, setting their Hamming distance to the total number of mismatched constituents and dependency arcs, respectively. At the inference phase, we use the CKY algorithm for constituency parsing and the Eisner algorithm for dependency parsing. 
    \item \textbf{Multi-task learning model (MTL)}.  Similar to \citet{zhou-zhao-2019-head}, we use the MTL framework at the training phase. The loss of MTL is the sum of losses for SEP. We use the head-binarization since preliminary experiments show that is superior to left- and right-binarization. Meanwhile, we apply the Eisner-Satta algorithm to perform joint parsing at the inference phase and ensure the compatibility of parsing results.

\end{itemize}

\paragraph{Hyper-parameters.} 
We employ \textsl{bert-large-cased} for English, \textsl{bert-base-chinese} for Chinese, and \textsl{bert-base-multilingual-cased} for SPMRL. We mainly adopt the same hyper-parameters of parsers from \citet{zhang-etal-2020-efficient, zhang-etal-2020-fast}. For the second-order model, we set the dimensions of $r^{\mathrm{word}}_i/r^{\mathrm{span}}_{i,j}$ to 500. 
We report results averaged over 3 runs with different random seeds for all experiments.

\makeatletter
\def\xmidrule{%
\noalign{\vskip\aboverulesep}%
\multispan{6}{\leaders\hbox to 4pt{\hss\vrule\@height\cmidrulewidth\@width 2pt \relax\hss}\hfill\kern0pt}\cr%
\noalign{\vskip\belowrulesep}%
}
\makeatother

\begin{table}[tb!]
    \centering 
    \footnotesize
    \begin{tabular*}
        {\columnwidth}{@{\extracolsep{\fill}}l cc ccc}
        \toprule
        & \multicolumn{2}{c}{\bf Dependency} & \multicolumn{3}{c}{\bf Constituency}\\
        \cmidrule(lr){2-3}\cmidrule(lr){4-6}
        & UAS & LAS & P & R & F1  \\
        \midrule 
       M\&H21$^{\dagger}$ & $96.66$ & $95.01$ & $-$ & $-$ & $-$ \\
       Y\&T22$^{\dagger}$ & $\pmb{97.24}$ & $\pmb{95.73}$ & $\pmb{96.19}$ & $\pmb{95.83}$ & $\pmb{96.01}$ \\
       HPSG & $97.00$ & $95.43$ & $95.98$ & $95.70$ & $95.84$ \\
       F\&G22 & $96.97$ & $95.46$ & $-$ & $-$ & $95.23$ \\
       \xmidrule
       MTL & $97.14$ & $95.59$ & $96.14$ & $95.80$ & $95.97$  \\
       Joint2o & $97.17$ & $95.64$ & $95.95$ & $95.65$ & $95.80$ \\
       \bottomrule 
    \end{tabular*}
    \caption{Comparison with previous results on PTB-test. 
    HPSG: \citet{zhou-zhao-2019-head}.
    M\&H21: \citet{mohammadshahi-henderson-2021-recursive}. 
    F\&G22: \citet{fer-2022}.
    Y\&T22: \citet{yang-tu-2022-bottom, yang-tu-2022-headed}.
    $^\dagger$ indicates using only the constituency or dependency treebank.  }
    \label{exp:ptb}
\end{table}

\makeatletter
\def\xmidrule{%
\noalign{\vskip\aboverulesep}%
\multispan{6}{\leaders\hbox to 4pt{\hss\vrule\@height\cmidrulewidth\@width 2pt \relax\hss}\hfill\kern0pt}\cr%
\noalign{\vskip\belowrulesep}%
}
\makeatother

\begin{table}[tb!]
    \centering 
    \footnotesize
    \begin{tabular*}
        {\columnwidth}{@{\extracolsep{\fill}}l cc ccc}
        \toprule
        & \multicolumn{2}{c}{\bf Dependency} & \multicolumn{3}{c}{\bf Constituency}\\
        \cmidrule(lr){2-3}\cmidrule(lr){4-6}
        & UAS & LAS & P & R & F1  \\
        \midrule 
       Zhang+20$^\dagger$ & $91.71 $& $90.38$  &  $\pmb{91.00}$ & $90.40$ & $90.70$  \\
       \xmidrule
       MTL & $92.11$ & $90.86$ & $90.75$ & $\pmb{90.91}$ & $\pmb{90.83}$  \\
       Joint2o & $\pmb{92.21}$ & $\pmb{90.96}$ & $90.73$ & $90.76$ & $90.76$ \\       
       \midrule
       & \multicolumn{5}{c}{\emph{w/ gold POS tags}} \\ 
       Zhang+20$^\dagger$ & $92.43$ & $92.04$ & $93.00$ & $92.98$ & $93.00$  \\
       M\&H21$^\dagger$ & $92.98$ & $91.18$ & $-$ & $-$ & $-$ \\
       Y\&T22$^\dagger$ & $93.33$ & $92.30$ & $-$ & $-$ & $-$ \\
       \xmidrule
       MTL & $93.16$ & $92.78$ & $\pmb{93.13}$ & $93.23$ & $93.18$ \\
       Joint2o & $\pmb{93.36}$ & $\pmb{92.97}$ & $\pmb{93.13}$ & $\pmb{93.25}$ & $\pmb{93.19}$ \\
       \bottomrule 
    \end{tabular*}
    \caption{Comparison with previous results on CTB-test. 
    Zhang+20: \citet{zhang-etal-2020-efficient, zhang-etal-2020-fast}. 
    M\&H21: \citet{mohammadshahi-henderson-2021-recursive}. 
    Y\&T22: \citet{yang-tu-2022-bottom}.
    $^\dagger$ indicates using only the constituency or dependency treebank.  
    }
    \label{exp:ctb}
\end{table}

\subsection{Main Results}

\begin{table*}[tb!]
    \centering 
    \setlength{\tabcolsep}{0.34em}
    \begin{tabular*}
        {\textwidth}{l cc cc cc cc cc cc}
        \toprule
        & \multicolumn{2}{c}{\footnotesize {\bf French}} & \multicolumn{2}{c}{\footnotesize {\bf Hebrew}} & \multicolumn{2}{c}{\footnotesize {\bf Korean}} & \multicolumn{2}{c}{\footnotesize {\bf Polish}} & \multicolumn{2}{c}{\footnotesize {\bf Swedish}} & \multicolumn{2}{c}{\footnotesize {\it Average}} \\
        \cmidrule(lr){2-3}\cmidrule(lr){4-5}\cmidrule(lr){6-7}\cmidrule(lr){8-9}\cmidrule(lr){10-11}\cmidrule(lr){12-13}
        & LAS & F1& LAS & F1& LAS & F1& LAS & F1& LAS & F1& LAS & F1 \\
        \midrule 

        \citet{kitaev-etal-2019-multilingual}$^\dagger$ & $-$ & $87.42$ & $-$ & $92.99$ & $-$ & $88.80$ & $-$ & $96.36$ & $-$ & $88.86$ & $-$ & $90.89$\\

        \citet{nguyen-etal-2020-efficient}$^\dagger$ & $-$ & $86.69$ & $-$ & $\pmb{93.67}$ & $-$ & $88.71$ & $-$ & $96.14$ & $-$ & $ 89.10$ & $-$ & $90.86$ \\

        \citet{yang-tu-2023-dont}$^\dagger$ & $-$ & $\pmb{87.89}$ & $-$ & $-$ & $-$ & $89.31$ & $-$ & $96.18$ & $-$ & $ - $ & $-$ & $-$ \\

        \citet{strzyz-etal-2019-sequence} & $83.85$ & $81.33$ & $74.94$ & $91.83$ & $85.93$ & $83.39$ & $85.86$ & $93.36$ & $79.77$ & $86.53$ & $82.07$ & $87.29$ \\

        \citet{li-zhou-2022-hpsg} & $88.53$ & $87.27$ & $85.21$ & $93.17$ & $\pmb{90.73}$ & $89.53$ & $\pmb{91.37}$ & $\pmb{96.55}$ & $87.69$ & $89.52$ & $87.60$ & $90.56$ \\
        
        \midrule
       MTL &$ 89.42 $&$ 87.58 $&$ 85.77 $&$ 93.09 $&$ 89.93 $&$ \pmb{89.46} $&$ 90.83 $&$ 96.33 $&$ 88.11 $&$ \pmb{89.73} $&$ 88.81 $&$ \pmb{91.24}$ \\ 
       
       Joint2o &$ \pmb{89.52} $&$ 87.48 $&$ \pmb{85.83} $&$ 92.96 $&$ 89.83 $&$ 89.21 $&$ 91.15 $&$ 96.37 $&$ \pmb{88.38} $&$ 89.69 $&$ \pmb{88.94} $&$ 91.14$ \\ 

       \bottomrule 
    \end{tabular*}
    \caption{Comparison with previous results on SPMRL-test. $^\dagger$ indicates using only the constituency or dependency treebank. Note that we only compare with the results of the HPSG parser in \citet{li-zhou-2022-hpsg}. 
    }
    \label{exp:ml}
\end{table*}

Table \ref{exp:cmp_ml} presents the results of our model study on multilingual test sets. 
On average, compared to SEP, all joint models (including MTL, Joint1o, and Joint2o) showed a substantial improvement (>0.2) for dependency parsing; and only MTL showed little impact on constituency parsing, where Joint1/2o instead slightly decreased the performance. In particular, Polish is a counterexample, with a trend just the opposite of the others. 

Between the joint models, we can see that Joint1o performs close to MTL in dependency parsing but lower in constituency parsing. Meanwhile, Joint2o incorporating higher-order features outperformed both Joint1o and MTL in dependency parsing. Even though Joint2o is higher than the first-order counterpart on constituency parsing, it is still lower than MTL. 

The above findings bring three insights: 1) joint parsing at the inference phase can indeed help dependency parsing and has little impact on constituency parsing; 2) further joint modeling at the training phase does not improve performance in first-order cases, 3) high-order modeling leads to tiny but steady improvements on both constituency and dependency parsing.

\subsection{Comparison with Previous Works}
Table \ref{exp:ptb}, \ref{exp:ctb} and \ref{exp:ml} compare our MTL and Joint2o with the existing state-of-the-art of both the separate and joint parsing on test sets. 
The performance gap between our parsers and recent state-of-the-art parsers is negligibly small.

On CTB, we re-run the code released by \citet{zhang-etal-2020-efficient, zhang-etal-2020-fast}, which provides the two strong TreeCRF parsers for c- and d-trees, to reproduce their results. We also note that reporting the results of using gold Part-Of-Speech (POS) tags for CTB is a more prevalent practice \citep{zhou-zhao-2019-head, mohammadshahi-henderson-2021-recursive, fer-2022, yang-tu-2022-headed}. Therefore, for comprehensive comparisons, we also conduct experiments with gold POS tags, by adding POS tag embeddings element-wisely to the hidden representations from the encoder.

\subsection{Versus the Pipeline Method } As discussed in Section \ref{sec:l-tree-repr}, we recognize that our model may result in labeling incompatibility. To verify whether labeling incompatibility affects parsing performance, we conduct experiments on the PTB/CTB using the pipeline method, first applying constituent parsing and then obtaining d-trees via const-to-dep conversion. 
Please note that the conversion processes utilized for PTB/CTB do not require function tags and thus we can directly apply them to the parsed c-trees.

\begin{table}[tb!]
    \centering 
    \small
    \begin{tabular*}
    {\columnwidth}{@{\extracolsep{\fill}}l ll ll}
        \toprule
        & \multicolumn{2}{c}{PTB} & \multicolumn{2}{c}{CTB} \\
        \cmidrule(lr){2-3}\cmidrule(lr){4-5}
        & {UAS} & {LAS} & {UAS} & {LAS} \\
        \midrule
        Pipeline  & $96.50$ & $95.37$ & $92.72$ & $92.14$ \\
        Joint2o & $\pmb{97.17}$ & $\pmb{95.64}$ & $\pmb{93.36}$ & $\pmb{92.97}$  \\
       \bottomrule
    \end{tabular*}
    \caption{Comparison with pipeline method on PTB/CTB-test.}
    \label{exp:cmp-with-pipeline}
\end{table}

Table \ref{exp:cmp-with-pipeline} shows that the performances of the pipeline method are inferior to that of Joint2o, especially on the CTB. We believe this is probably because such a conversion process suffers from error propagation, although compatibility is guaranteed.

Notably, when evaluated on the relatively high-performing PTB, the pipeline method exhibits a smaller performance drop on LAS (0.27) compared to UAS (0.67). This suggests that enforcing the labeling compatibility may lead to a better label prediction and therefore become a promising extension.  
While it seems impossible to extend strict labeling compatibility in our model, we can try to achieve approximate compatibility. By simplifying the label determination rules to be similar to those for finding head words, we can count such rules in the training set and adopt them to constrain the label prediction. We leave such extension of compatibility as future work.

\begin{table}[tb!]
    \centering 
    \begin{tabular*}
    {\columnwidth}{@{\extracolsep{\fill}}l c}
        \toprule
        Model & Sents/sec \\
        \midrule 
        HPSG  & $122.01$ \\
        MTL {\scriptsize w/ CKY+Eisner} & $299.31$  \\
        MTL {\scriptsize w/ Eisner-Satta} & $311.80$  \\
        Joint2o {\scriptsize w/ Eisner-Satta2o} & $305.79$ \\
        \bottomrule 
    \end{tabular*}
    \caption{Speed comparison on PTB-test. }
    \label{exp:speed}
\end{table}

\subsection{Speed Comparison}
Table \ref{exp:speed} compares different parsing models in terms of parsing speed, including HPSG from \citet{zhou-zhao-2019-head} with $O(n^5)$ parsing algorithm in Cython implementation, MTL with batchified CKY and Eisner from \citet{zhang-etal-2020-fast, zhang-etal-2020-efficient}, and MTL/Joint2o with our batchified Eisner-Satta and its 2nd-order extension respectively. Our models are run on a machine with Intel Xeon Gold 6248R CPU and NVIDIA A100 40GB GPU. We set the batch size to 100 sentences and report the average time over ten runs. 

We can observe that models with batchified algorithms are roughly 2.5$\times$ faster than HPSG19. Also, the speeds of ``CKY+Eisner" and ``Eisner-Satta(2o)" have similar speeds, probably because their time complexity after batchifying are all $O(n)$ on GPUs.

\subsection{Analysis}

In the previous experiments, we only used attachment scores and F1 scores to evaluate the performance of parsing, which assess the accuracy of dependencies and constituents respectively. However, upon analyzing the results in Table \ref{exp:cmp_ml}, we observe that the differences between the models are relatively small, especially for constituency parsing. For more detailed analyses, we evaluate the parsing results from other perspectives.

\paragraph{Complete match of syntactic trees. }
To facilitate a more evident comparison between different models, following \citet{zhang-etal-2019-empirical}, we adopt a more challenging metric known as the sentence-level labeled complete match (LCM). Specifically, we evaluated the LCM of the predicted c-trees, d-trees, and their combination, denoted as LCM$_{con}$/LCM$_{dep}$/LCM$_{con+dep}$ respectively. Please note that an instance where the LCM$_{con+dep}$ is true only if both the LCM$_{con}$ and LCM$_{dep}$ are true.

\begin{table}[tb!]
\centering
\small
\begin{tabular*}
{\columnwidth}{@{\extracolsep{\fill}}l ccc}
\toprule
& LCM$_{con}$ & LCM$_{dep}$ & LCM$_{con+dep}$ \\
\midrule
\multicolumn{4}{c}{\bf PTB} \\
SEP & $55.25_{\pm 0.6}$ & $54.66_{\pm 0.5}$ & $43.27_{\pm 0.6}$ \\
MTL & $\pmb{55.45}_{\pm 0.6}$ & $55.74_{\pm 0.4}$ & $46.74_{\pm 0.6}$ \\
Joint1o & $55.02_{\pm 0.4}$ & $55.84_{\pm 0.4}$ & $46.88_{\pm 0.5}$ \\
Joint2o & $55.37_{\pm 0.5}$ & $\pmb{55.95}_{\pm 0.5}$ & $\pmb{46.97}_{\pm 0.5}$ \\
\midrule
\multicolumn{4}{c}{\bf CTB} \\
SEP & $29.37_{\pm 0.4}$ & $41.53_{\pm 0.4}$ & $26.03_{\pm 0.5}$ \\
MTL & $\pmb{29.39}_{\pm0.3}$ & $43.08_{\pm 0.4}$ & $27.99_{\pm 0.3}$ \\
Joint1o & $29.16_{\pm 0.4}$ & $43.14_{\pm 0.3}$ & $28.25_{\pm 0.4}$ \\
Joint2o & $29.11_{\pm 0.3}$ & $\pmb{43.36}_{\pm 0.4}$ & $\pmb{28.33}_{\pm 0.5}$ \\
\bottomrule
\end{tabular*}
\caption{Complete matching on PTB/CTB-test.}
\label{exp:cm}
\end{table}

For LCM$_{con}$ and LCM$_{dep}$, their trends are quite similar to the model study in Table \ref{exp:cmp_ml}. For LCM$_{con+dep}$, we can clearly see that all the joint models have a significant improvement compared to SEP (>3 on PTB), suggesting that joint parsing c- and d-trees can lead to stronger compatibility.

Among joint parsing models, Joint1o has a slight improvement in LCM$_{con+dep}$ compared to MTL, and there is a further gain on Joint2o. This trend on both PTB and CTB indicates that joint parsing further at the training phase and the second-order extension both help the compatibility.

\paragraph{Fine-grained analysis. }
\citet{yang-tu-2022-headed} shows that different tree modelings may influence the performance on different lengths of sequence, span, and dependency. 
To investigate this, we follow \citet{yang-tu-2022-headed} and plot LAS/F1-score as functions of the sequence lengths, constituent widths, and dependency lengths on CTB-test\footnote{We also conduct the same experiment on PTB-test and find a similar trend. However, all these models perform well on PTB and the margins between them are relatively small, resulting in fewer distinctions. 
}. The width of a constituent from $w_i$ to $w_j$ is equal to $d:= |i - j|$.

\begin{figure}[tb!]
    \centering
    \subfigure[]{
        \label{fig:c1}    
        \includegraphics[scale=0.31,trim=8 0 5 0,clip]{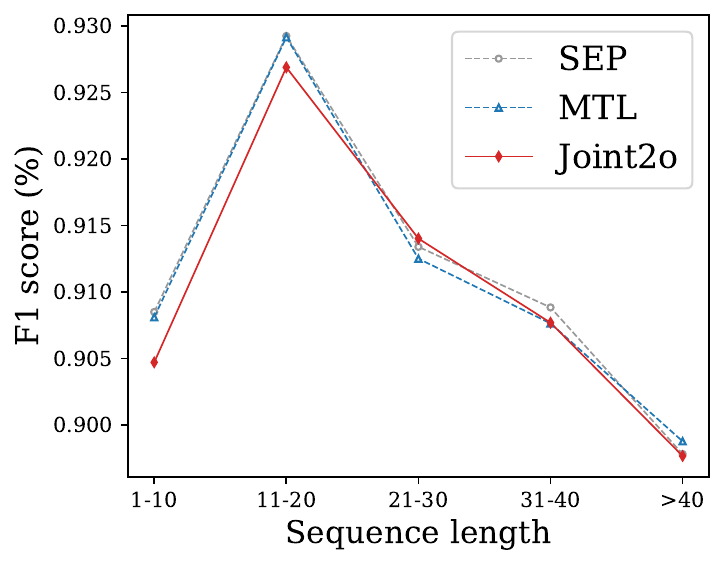}
    }
    \hspace{-3mm}
    \subfigure[]{
        \label{fig:c2}    
        \includegraphics[scale=0.31,trim=8 0 5 0,clip]{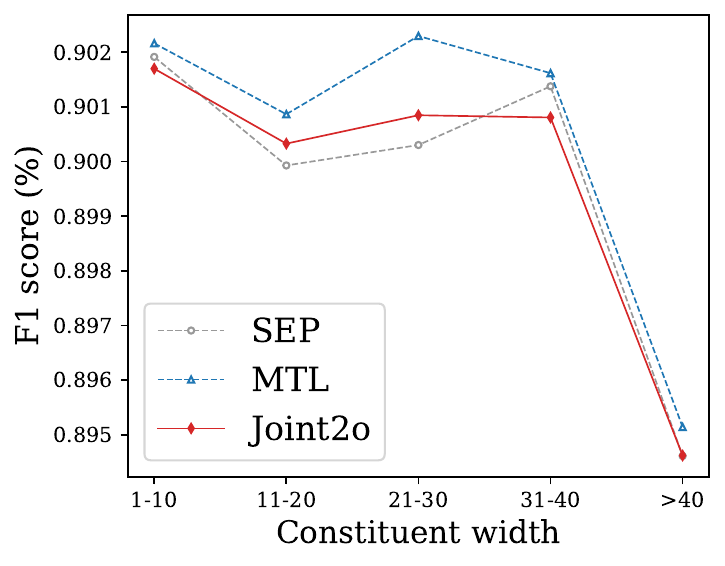}
    }
    \\
    \subfigure[]{
        \label{fig:d1}    
        \includegraphics[scale=0.31,trim=8 0 5 0,clip]{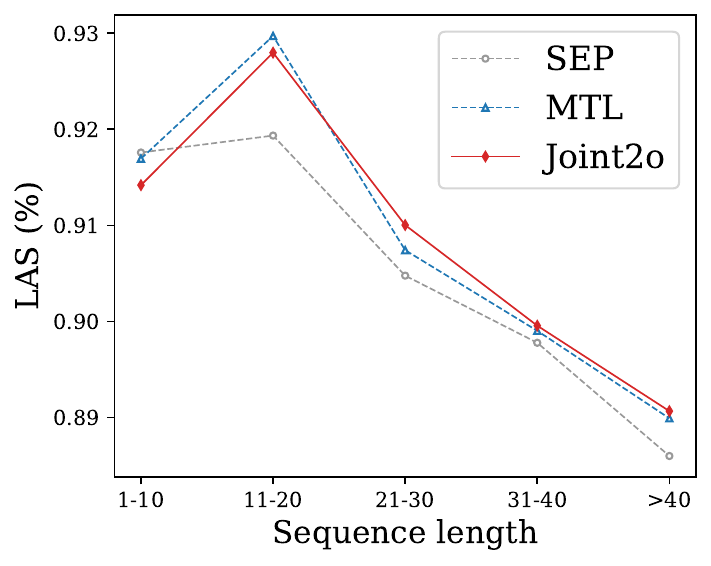}
    }
    \hspace{-3mm}
    \subfigure[]{
        \label{fig:d2}    
        \includegraphics[scale=0.32,trim=8 0 5 0,clip]{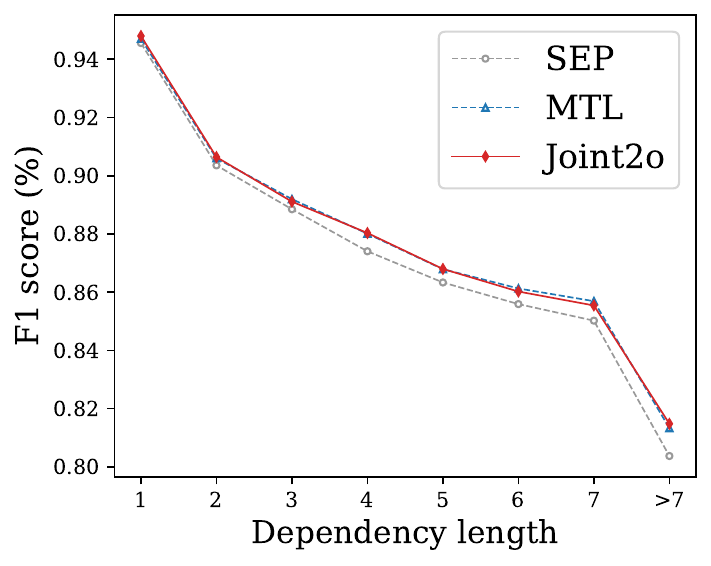}
    }
  \caption{Fine-grained analysis on CTB-test. }
  \label{fig:con&dep-ana}
\end{figure}

Figure \ref{fig:c1} and \ref{fig:c2} show the F1 scores for different sequence lengths and constituent widths. Notably, MTL and SEP perform similarly on sentences of different lengths, while Joint2o is slightly weaker on short sentences (<20). For constituent widths, MTL and Joint2o both have a small increase compared to SEP on short and medium constituents (<30), however, Joint2o is lower than the other two on long constituents (>30).

From Figure \ref{fig:d1} and \ref{fig:d2}, we can see that MTL and Joint2o both have a steady improvement over SEP on almost all sentence lengths and dependency lengths. Meanwhile, Joint2o performs slightly better than MTL on longer sentences (>20).

\section{Conclusions}
This paper revisits the topic of joint c-tree and d-tree parsing. 
Compared with previous works, we further explore joint parsing at the training phase, thanks to the improved efficiency from the Eisner-Satta \cite{eisner-satta-1999-efficient} decoding algorithm.  We also design second-order scoring components for promoting interaction between constituents and dependencies. 

Our experiments encompass benchmark datasets in seven languages, yielding the following key findings. 
\begin{enumerate}
\item The Eisner-Satta algorithm leads to about 2.5$\times$ speed-up, compared with the decoding algorithm proposed by \citet{zhou-zhao-2019-head}. 
\item Compared with separate modeling, joint parsing only at the inference phase, i.e., the MTL approach of \citet{zhou-zhao-2019-head},  leads to modest performance boost on d-tree parsing, and has little impact on c-tree parsing performance. 
\item Joint modeling at both training and inference phases does not further improve performance, compared with joint modeling only at the inference phase. 

\item High-order joint modeling leads to modest performance improvement on both d-tree and c-tree parsing, compared with the first-order counterpart. 
\item Detailed analysis shows that joint parsing significantly improves the complete matching ratio for the combination of c- and d-trees.
\end{enumerate}

\section*{Acknowledgements}
We thank all the anonymous reviewers for their valuable comments. We also thank Yu Zhang for his well-designed package Supar\footnote{https://github.com/yzhangcs/parser/tree/main}, from which we borrow the batchified version of the first-order Eisner-Satta algorithms in our work. This work was supported by National Natural Science Foundation of China (Grant No. 62176173 and 62336006), and a Project Funded
by the Priority Academic Program Development (PAPD) of Jiangsu Higher Education Institutions.

\section*{Bibliographical References}
\label{sec:reference}
\bibliographystyle{lrec-coling2024-natbib}
\bibliography{lrec-coling2024}

\section*{Language Resource References}
\label{lr:ref}
\bibliographystylelanguageresource{lrec-coling2024-natbib}
\bibliographylanguageresource{languageresource}

\end{document}